%% file: main.tex
\pdfoutput=1

\documentclass[11pt]{article}

\usepackage{acl}

\usepackage{times}
\usepackage{latexsym}

\usepackage[T1]{fontenc}

\usepackage[utf8]{inputenc}

\usepackage{microtype}

\usepackage{inconsolata}

\usepackage{listings}
\usepackage{caption}
\usepackage{graphicx}
\usepackage{makecell}
\usepackage{subcaption}
\usepackage{microtype}
\usepackage{multirow}
\usepackage{bm}
\usepackage{tabularx}
\usepackage{ltablex}
\usepackage{longtable}
\usepackage{colortbl}
\usepackage{supertabular}
\usepackage{xltabular}
\usepackage{adjustbox}
\usepackage{lineno}
\usepackage{array}
\usepackage{booktabs}
\usepackage{hyperref}
\usepackage{amsmath}
\usepackage{amsfonts}
\usepackage{amssymb}
\usepackage{makecell}
\usepackage{paralist}

\definecolor{yamlKeyColor}{RGB}{0, 119, 170}    
\definecolor{yamlStringColor}{RGB}{126, 49, 170} 
\definecolor{yamlCommentColor}{RGB}{112, 128, 144} 
\definecolor{yamlBackground}{RGB}{250, 250, 250} 

\lstdefinestyle{yamlstyle}{
    language={},
    basicstyle=\ttfamily\small,
    numbers=none,
    backgroundcolor=\color{yamlBackground},
    showspaces=false,
    showstringspaces=false,
    showtabs=false,
    frame=single,
    framesep=5pt,
    framerule=0.4pt,
    tabsize=2,
    captionpos=b,
    breaklines=true,
    breakatwhitespace=true,
    breakautoindent=true,
    linewidth=\textwidth,
    commentstyle=\color{yamlCommentColor}\itshape,
    keywordstyle=\color{yamlKeyColor},
    stringstyle=\color{yamlStringColor},
    moredelim=[l][\color{yamlKeyColor}]{-},
    moredelim=[l][\color{yamlKeyColor}]{:}
}

\title{MoD: A Distribution-Based Approach for Merging Large Language Models}

\author{Quy-Anh Dang$^{*1,2}$, Chris Ngo$^{2}$ \\
$^1$VNU University of Science, Vietnam \\
$^2$Knovel Engineering Lab, Singapore \\
\texttt{dangquyanh150101@gmail.com, chris.ngo@knoveleng.com}
}

\begin{document}
\maketitle

\begin{abstract}
Large language models (LLMs) have enabled the development of numerous specialized, task-specific variants. However, the maintenance and deployment of these individual models present substantial challenges in terms of resource utilization and operational efficiency.
In this work, we propose the \textit{Mixture of Distributions (MoD)} framework, a novel approach for merging LLMs that operates directly on their output probability distributions, rather than on model weights. Unlike traditional weight-averaging methods, MoD effectively preserves the specialized capabilities of individual models while enabling efficient knowledge sharing across tasks. Through extensive experimentation on mathematical reasoning benchmarks using Qwen2.5 models, we demonstrate that MoD significantly outperforms existing model merging techniques across multiple benchmarks. All code, data, and experimental materials are published at \url{https://github.com/knovel-eng/mod}.
\end{abstract}

\def\thefootnote{(*)}\footnotetext{English Name is Andrew Dang}\def\thefootnote{\arabic{footnote}}

\thispagestyle{plain}
\pagestyle{plain}

\begin{figure*}[ht]
\centering
\begin{subfigure}{\textwidth}
\centering
\includegraphics[width=\textwidth]{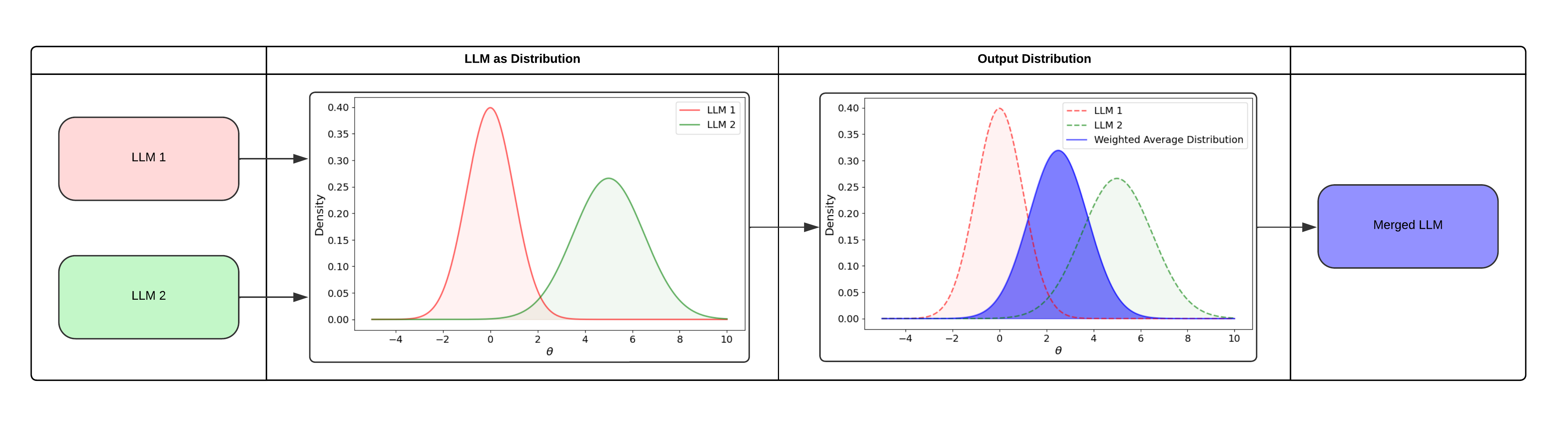}
\caption{Weighted Average Method}
\label{fig:wa-pipeline}
\end{subfigure}

\vspace{1em} 
\begin{subfigure}{\textwidth}
    \centering
    \includegraphics[width=\textwidth]{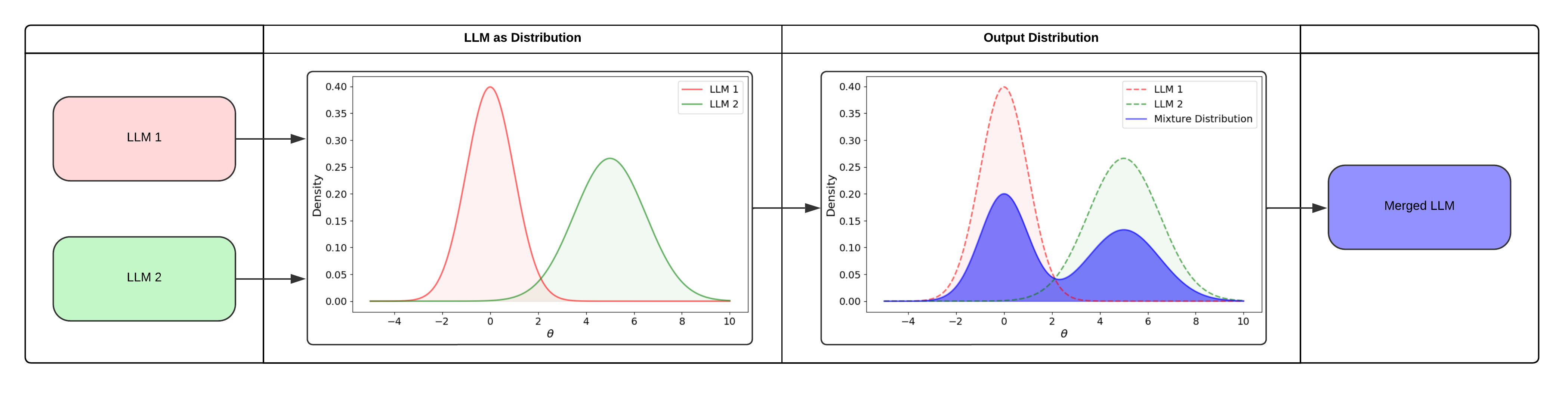}
    \caption{MoD (Our Method)}
    \label{fig:main-pipeline}
\end{subfigure}
\caption{Comparison of our \textbf{MoD} method with the Weighted Average method. While weighted averaging methods for merging LLMs often produce new distributions that alter the characteristics of the original models (see Fig. \ref{fig:wa-pipeline}), our MoD approach effectively preserves key density structures, accurately maintaining peak densities at $\theta = 0$ and $\theta = 5$ (see Fig. \ref{fig:main-pipeline}).}
\label{fig:pipeline}

\end{figure*}

\section{Introduction}
In the past year, we have witnessed significant advancements in open-source large language models (LLMs), many of which are available on the Hugging Face model hub \citep{wolf-etal-2020-transformers}. These models are trained on datasets containing trillions of tokens and range from 1 to 70 billion parameters \citep{minaee2024largelanguagemodelssurvey, zhang2024tinyllamaopensourcesmalllanguage}. The diversity of open-source checkpoints is remarkable, with a broad classification into pretrained models \citep{9134370} and models fine-tuned for instruction-following across a range of domains, such as coding \citep{rozière2024codellamaopenfoundation} and medical applications \citep{wu2023pmcllamabuildingopensourcelanguage}. However, fine-tuning separate models for each specific task poses two key challenges:

\begin{enumerate}
    \item Each task-specific model must be stored and deployed independently, leading to increased storage and deployment costs.
    \item Independently trained models are unable to share insights across tasks, limiting their ability to enhance task-specific performance or generalize to other domains \citep{sanh2022multitaskpromptedtrainingenables, rame2022recycling, yu2024language}.
\end{enumerate}

Training these models from scratch is resource-intensive, as illustrated by the Mistral-7B model, which incurred costs between 2 to 3 million USD \citep{jiang2023mistral7b}. Further fine-tuning of pretrained models often results in catastrophic forgetting \citep{10444954}, where the model's original generalization capabilities degrade, impairing its performance across multiple tasks \citep{cheng2024adapting, wu-etal-2024-llama}. Moreover, aligning models to human preferences demands substantial effort and data collection, making it impractical for most research teams to replicate \citep{aligning_llm_human, rafailov2023direct}.

These challenges bring into focus the critical question of how to best utilize existing pretrained checkpoints for research and practical applications. In this context, model merging has emerged as a promising approach, combining parameters from multiple task-specific models into a single, unified model. This technique enables multitask and continual learning while minimizing catastrophic forgetting, all without the steep costs of training models from scratch \citep{yadav2023tiesmerging}.

In this paper, we present a novel method called Mixture of Distributions (MoD) for merging LLMs, which extends previous approaches by introducing a probabilistic framework that optimizes the balance between task specialization and generalization. MoD merges models by constructing a mixture distribution over their parameters, preserving the strengths of each model and reducing the risk of catastrophic forgetting. Through experiments with Qwen2.5-1.5B and Qwen2.5-7B models \citep{qwen2.5, qwen2}, we demonstrate that MoD significantly surpasses traditional merging techniques, especially in multitask scenarios involving mathematics tasks.

Our main contributions are:

\begin{itemize}
    \item We propose Mixture of Distributions (MoD), a new and efficient method for merging LLMs, which outperforms existing methods.
    \item We provide extensive experimental results using Qwen2.5 models on tasks in the math domain, highlighting superior performance.
    \item We release all code to facilitate future research in this area\footnote{\url{https://github.com/knovel-eng/mod}}.
\end{itemize}

\section{Related Work}
\paragraph{Model Merging} Recent advances in large language models (LLMs) have highlighted model merging as a crucial strategy for combining the capabilities of multiple models into a unified system \citep{ainsworth2023gitrebasinmergingmodels, goddard2024arceesmergekittoolkitmerging, labrak-etal-2024-biomistral}. This approach has gained prominence for its ability to enhance multitask performance and enable continual learning without requiring costly retraining procedures. Initial investigations in this domain explored weight averaging techniques, which directly combined parameters of models sharing identical architectures and initializations \citep{NEURIPS2022_70c26937, NEURIPS2018_be3087e7}. While these methods demonstrated promising results, they revealed significant limitations when applied to models trained on heterogeneous tasks or initialized differently, prompting the development of more sophisticated approaches.

\paragraph{Merging Techniques} The theoretical foundation for many modern merging approaches stems from Linear Mode Connectivity (LMC) \citep{pmlr-v119-frankle20a}, which demonstrates that models fine-tuned from a common pretrained checkpoint often permit linear interpolation while maintaining performance integrity \citep{nagarajan2021uniformconvergenceunableexplain, neyshabur2021transferredtransferlearning}. This insight has led to the development of several practical methodologies. Model Soups \citep{pmlr-v162-wortsman22a} and weight averaging techniques \citep{NEURIPS2022_70c26937, NEURIPS2018_be3087e7} offer elegant solutions for merging models with shared initialization. Task Arithmetic \citep{ilharco2023editing} extends this framework by introducing task vectors, demonstrating that arithmetic operations on the differences between fine-tuned models and their base model yield semantically meaningful results. More recent approaches, including Trim, Elect Sign \& Merge (TIES merging) \citep{yadav2023tiesmerging}, Model Breadcrumbs \citep{davari2024model}, and Drop And REscale (DARE) \citep{yu2024language}, have introduced sophisticated methods for sparsifying and combining task vectors, enabling the integration of multiple models while preserving their individual capabilities. The application of Spherical Linear intERPolation (SLERP) \citep{shoemake1985animating} represents a significant advancement over simple weight averaging, revealing that spherical paths often present lower loss barriers compared to direct linear interpolation.

The challenge of merging independently trained models with different initializations presents a more complex scenario. Git-Rebasin \citep{ainsworth2023gitrebasinmergingmodels} addresses this challenge by exploiting neural networks' permutation symmetry, enabling the alignment of neurons across independently trained models to facilitate effective merging. Complementary approaches such as Optimizing Mode Connectivity via Neuron Alignment \citep{NEURIPS2020_aecad423} and Optimal Transport Fusion (OTFusion) \citep{imfeld2024transformerfusion} have further developed this concept, demonstrating enhanced capabilities in reducing interpolation barriers between models with distinct random initializations.

Recent research has pushed the boundaries of model merging by exploring the integration of models with heterogeneous architectures. The Composition to Augment Language Models (CALM) approach \citep{bansal2024llmaugmentedllmsexpanding} leverages cross-attention mechanisms to integrate models with diverse neural architectures, marking a significant advancement in the field. Similarly, the FUSELLM framework \citep{wan2024knowledge} focuses on aligning probabilistic distributions across different language models, facilitating the fusion of models with varying output characteristics. While these methods incur higher computational costs and may require additional pretraining, they represent important progress toward creating more versatile and adaptable models.

In this paper, we introduce Mixture of Distributions (MoD), a novel approach that shifts the paradigm from weight interpolation to probabilistic output combination. Our method leverages the probability density functions of large language models, enabling a more nuanced integration that preserves the distinctive strengths of each model. The following sections detail the methodology of MoD (Section \ref{Methodology}), present our experimental validation (Section \ref{Experiments}), provide conclusions (Section \ref{Conclusion}), and discuss limitations and future research directions (Section \ref{Limitations-and-Future-Work}).

\section{Methodology} \label{Methodology}

In this section, we present the \textbf{Mixture of Distributions (MoD)} method for merging large language models (LLMs) through direct combination of their output probability distributions. By operating in the distribution space rather than interpolating model weights, MoD effectively addresses key challenges, particularly the distortion of density functions commonly observed in traditional weight-based approaches. We establish the mathematical framework, present the underlying motivation, and detail the implementation of our approach.

\subsection{Notation and Symbols} \label{Notation and Symbols}
Let $ \mathbf{\theta}_1 $ and $ \mathbf{\theta}_2 $ denote the parameter sets (weights) of two large language models (LLMs), where each $ \theta_i $ follows a multivariate normal distribution, $ \theta_i \sim \mathcal{N}(\mu_i, \Sigma_i) $. Given a sequence of input tokens $ x $, let $ p_{\mathbf{\theta}_1}(x) $ and $ p_{\mathbf{\theta}_2}(x) $ represent the probability density functions (PDFs) of the two models evaluated at $ x $. Our objective is to derive a unified output distribution $ p_{\theta}(x) $ that preserves the essential characteristics of both original models while providing a coherent merged representation.

Traditional weight-based merging approaches \citep{NEURIPS2022_70c26937, NEURIPS2018_be3087e7} compute the merged model's parameters through linear combination:
\[
\theta = \alpha \mathbf{\theta}_1 + (1 - \alpha) \mathbf{\theta}_2
\]
where $ \alpha \in [0, 1] $. However, this approach frequently distorts the original probability distributions.

\subsection{Motivation} \label{Motivation}
\begin{figure}[h]
\centering
\includegraphics[width=0.5\textwidth]{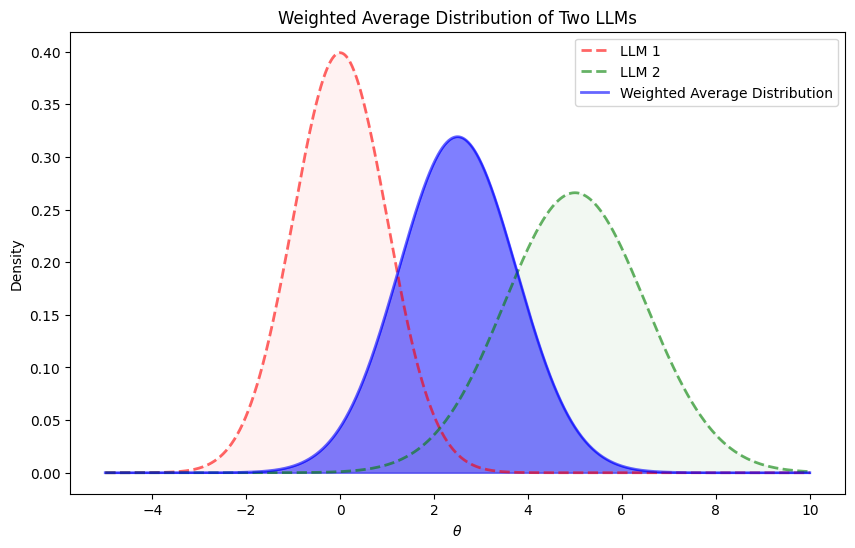}
\caption{Distribution distortion in weighted averaging methods, demonstrating failure to preserve maximum density at $\theta = 0$ despite high density in the \textcolor{red}{red} distribution.}
\label{fig:weighted-avg-distribution}
\end{figure}

The weighted averaging \citep{NEURIPS2022_70c26937, NEURIPS2018_be3087e7} of model parameters introduces significant distributional shifts, resulting in unintended density alterations. Specifically, linear interpolation between $ \mathbf{\theta}_1 $ and $ \mathbf{\theta}_2 $ can lead to high density assignments at points $ x $ where neither $ p_{\mathbf{\theta}_1}(x) $ nor $ p_{\mathbf{\theta}_2}(x) $ initially exhibited significant probability mass. This phenomenon results in poor generalization performance on downstream tasks, as illustrated in Figure \ref{fig:weighted-avg-distribution}.

To address these limitations, we propose a distribution-centric merging method that operates directly on the models' output PDFs. The Mixture of Distributions (MoD) method ensures preservation of both models' probabilistic properties while maintaining their fundamental density structures.

\subsection{Mixture of Distributions (MoD)} \label{Mixture of Distributions}
The MoD framework directly combines the output distributions of the constituent models. Rather than merging parameters, we define the unified output distribution as a weighted combination of probability densities:
\[
p_{\theta}(x) = \alpha p_{\mathbf{\theta}_1}(x) + (1 - \alpha) p_{\mathbf{\theta}_2}(x)
\]
where $ \alpha \in [0, 1] $. Here, $ p_{\mathbf{\theta}_1}(x) $ and $ p_{\mathbf{\theta}_2}(x) $ represent the probability density functions of models 1 and 2 at point $ x $, with $ \alpha $ and $ (1 - \alpha) $ serving as mixture weights.

\paragraph{Solving for Mixture Weights}
The determination of optimal mixture weights requires solving:
\begin{equation}
    \begin{aligned}
        f: \mathbb{R}^{n} \times \mathbb{R}^{n} \to \mathbb{R}^{n} \\
        \theta = f(\mathbf{\theta}_1, \mathbf{\theta}_2)
    \end{aligned}
\end{equation}
where $ f $ represents the mapping function that identifies appropriate mixture weights while maintaining distributional dimensionality. We approach this through quantile function analysis. The quantile function $ Q(p) $ identifies the value $ \theta_{\text{specific}} $ such that:
\[
Q(p) = \inf \{ \theta_{\text{specific}} \in \mathbb{R} : P(\mathbf{\theta} \leq \theta_{\text{specific}}) = p \}
\]
where $ P(\mathbf{\theta} \leq \theta_{\text{specific}}) $ represents the cumulative distribution function (CDF) of the mixture distribution, and $\theta_{\text{specific}}$ denotes a specific value in the distribution of $\mathbf{\theta}$. To address the computational complexity of quantile function optimization, we employ a threshold-based approach. We normalize $ \theta_1 $ within $ [0, 1] $ as $\theta_{1-\text{normalize}}$ and choose $\alpha$ as a threshold that governs distributional contributions:
\[
\theta = 
\begin{cases}
    \theta_1, & \text{if } \theta_{1-\text{normalize}} < \alpha \\
    \theta_2, & \text{otherwise.}
\end{cases}
\]
This formulation ensures selective integration of significant distributional components.

\begin{figure}[h]
\centering
\includegraphics[width=0.5\textwidth]{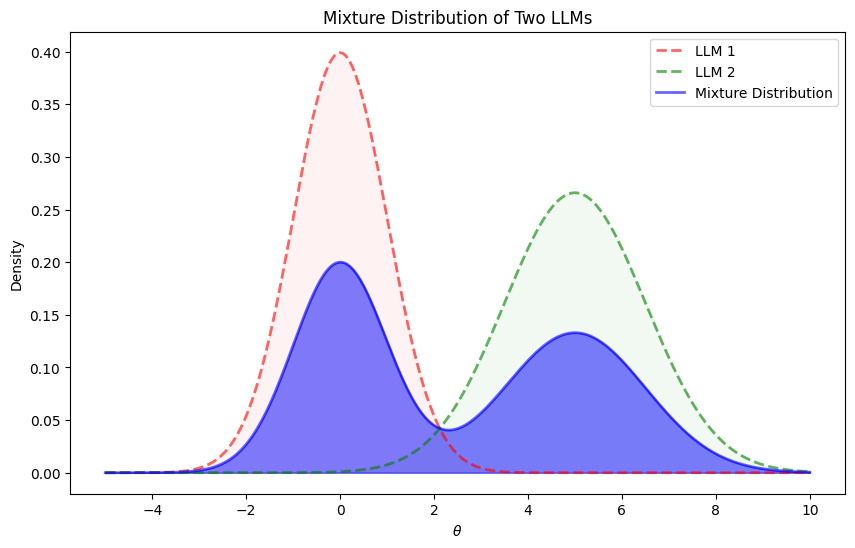}
\caption{MoD successfully preserves maximum density characteristics at $\theta = 0$, demonstrating effective distribution merging compared to traditional approaches.}
\label{fig:mixture-distribution}
\end{figure}

\paragraph{Maximizing Density at Key Points}
A core advantage of MoD is its preservation of original density structures while emphasizing critical distributional regions. Unlike weight-averaging methods, which often generate spurious density peaks, MoD maintains density characteristics at crucial points $ x $ through dynamic adjustment of mixture weights based on input sequences (Figure \ref{fig:mixture-distribution}).

\section{Experiments} \label{Experiments}

\begin{table}[h]
\centering
\caption{Datasets and Number of Samples for Evaluation}
\label{table:datasets}
\begin{tabular}{@{}l | r@{}}
\toprule
\textbf{Dataset} & \textbf{\#Num Samples} \\
\midrule
GSM8K & 1319 \\
MATH & 5000 \\
College Math & 2818 \\
SVAMP & 1000 \\
ASDiv & 2215 \\
MAWPS & 2065 \\
CARP En & 976 \\
GaoKao 2023 En & 385 \\
OlympiadBench & 675 \\
MMLU STEM & 3018 \\
AIME24 & 30 \\
AMC23 & 40 \\
\bottomrule
\end{tabular}
\end{table}

\paragraph{Datasets} 
To evaluate the performance and efficiency of our merged model, we focus on a range of mathematics-focused benchmark datasets. Specifically, we use twelve datasets representing diverse aspects of mathematical reasoning and problem-solving:
\begin{itemize}
    \item \textbf{GSM8K} \citep{cobbe2021gsm8k}: A linguistically diverse set of grade school-level math word problems created by human authors.
    \item \textbf{MATH} \citep{hendrycksmath2021}: A collection of problems from various mathematics competitions.
    \item \textbf{SVAMP} \citep{patel-etal-2021-nlp}: The Simple Variations on Arithmetic Math Problems dataset, designed to assess robustness in solving arithmetic word problems.
    \item \textbf{ASDiv} \citep{miao-etal-2020-diverse}: The Arithmetic Semantic Dataset with Diverse Variations, another benchmark for arithmetic problem-solving.
    \item \textbf{MAWPS} \citep{patel-etal-2021-nlp}: A dataset of arithmetic word problems from a variety of sources.
    \item \textbf{CARP En} \citep{zhang2023evaluating}: Contains computation-intensive algebra problems from school curricula.
    \item \textbf{GaoKao 2023 En} \cite{Zhang2023EvaluatingTP}: An English version of problems from the 2023 Chinese National College Entrance Examination (Gaokao).
    \item \textbf{OlympiadBench} \citep{he-etal-2024-olympiadbench}: A bilingual, multimodal benchmark for Olympiad-level mathematics and physics problems.
    \item \textbf{College Math} \citep{tang2024mathscale}: Math problems targeting the college level.
    \item \textbf{MMLU STEM} \citep{hendryckstest2021, hendrycks2021ethics}: The STEM subset of the Massive Multitask Language Understanding (MMLU) benchmark.
    \item \textbf{AIME24}\footnote{\url{https://huggingface.co/datasets/AI-MO/aimo-validation-aime}}: Problems from the 2024 American Invitational Mathematics Examination.
    \item \textbf{AMC23}\footnote{\url{https://huggingface.co/datasets/AI-MO/aimo-validation-amc}}: Problems from the 2023 American Mathematics Competitions.
\end{itemize}

The details of these datasets, including the number of samples, are summarized in Table \ref{table:datasets}.

\paragraph{Metrics} 
We report 5-shot pass@1 \citep{song2022comprehensivesurveyfewshotlearning, chen2021codex} performance for MMLU (STEM) and zero-shot pass@1 performance on the remaining benchmarks \citep{kojima2023largelanguagemodelszeroshot}. 

\paragraph{Baseline Models} 
\begin{table*}[htbp]
\centering
\setlength{\tabcolsep}{2.5pt}
\renewcommand{\arraystretch}{1.3}
\small
\caption{Performance comparison of different methods across mathematical benchmarks by merging Qwen2.5-1.5B-Instruct and Qwen2.5-1.5B-Math-Instruct}
\label{tab:method-comparison-1.5B}
\begin{tabular}{l|cccccccccccc}
\hline
\multirow{2}{*}{Method} & \multicolumn{12}{c}{Benchmarks} \\
\cline{2-13}
& GSM8K & MATH & \begin{tabular}[c]{@{}c@{}}College\\Math\end{tabular} & SVAMP & ASDiv & MAWPS & \begin{tabular}[c]{@{}c@{}}CARP\\En\end{tabular} & \begin{tabular}[c]{@{}c@{}}GaoKao\\2023 En\end{tabular} & \begin{tabular}[c]{@{}c@{}}Olympiad\\Bench\end{tabular} & \begin{tabular}[c]{@{}c@{}}MMLU\\STEM\end{tabular} & AIME24 & AMC23 \\
\hline
Linear & 39.3 & 11.7 & 9.5 & 61.1 & 64.1 & 76.1 & 23.6 & 14.3 & 3.1 & 34.8 & 0.0 & 2.5 \\
Task-Arithmetic & 47.2 & 27.9 & 18.5 & 74.3 & 79.6 & 85.5 & 40.4 & 28.1 & 7.0 & 19.4 & 0.0 & 17.5 \\
TIES & 16.5 & 12.0 & 9.5 & 46.4 & 50.9 & 54.1 & 18.8 & 10.9 & 2.4 & 5.8 & 0.0 & 0.0 \\
DARE & 51.5 & 22.1 & 13.9 & 64.3 & 69.8 & 78.8 & 34.2 & 21.8 & 6.1 & 45.1 & 3.3 & 10.0 \\
SLERP & 47.3 & 20.9 & 13.9 & 58.1 & 64.8 & 70.3 & 31.8 & 21.8 & 6.2 & 42.4 & 0.0 & 7.5 \\
MoD (Our) & \textbf{74.5} & \textbf{55.8} & \textbf{38.0} & \textbf{85.1} & \textbf{88.0} & \textbf{95.1} & \textbf{56.0} & \textbf{47.0} & \textbf{20.6} & \textbf{59.5} & \textbf{10.0} & \textbf{27.5} \\
\hline
\end{tabular}
\end{table*}

\begin{table*}[htbp]
\centering
\setlength{\tabcolsep}{2.5pt}
\renewcommand{\arraystretch}{1.3}
\small
\caption{Performance comparison of different methods across mathematical benchmarks by merging Qwen2.5-7B-Instruct and Qwen2.5-7B-Math-Instruct}
\label{tab:method-comparison-7B}
\begin{tabular}{l|cccccccccccc}
\hline
\multirow{2}{*}{Method} & \multicolumn{12}{c}{Benchmarks} \\
\cline{2-13}
& GSM8K & MATH & \begin{tabular}[c]{@{}c@{}}College\\Math\end{tabular} & SVAMP & ASDiv & MAWPS & \begin{tabular}[c]{@{}c@{}}CARP\\En\end{tabular} & \begin{tabular}[c]{@{}c@{}}GaoKao\\2023 En\end{tabular} & \begin{tabular}[c]{@{}c@{}}Olympiad\\Bench\end{tabular} & \begin{tabular}[c]{@{}c@{}}MMLU\\STEM\end{tabular} & AIME24 & AMC23 \\
\hline
Linear & 91.9 & 70.7 & 45.6 & 92.7 & 95.1 & 97.9 & 58.8 & 62.1 & 35.0 & 56.9 & \textbf{13.3} & \textbf{47.5} \\
Task-Arithmetic & 72.5 & 40.8 & 24.0 & 84.9 & 89.8 & 92.8 & 47.2 & 37.7 & 12.0 & 32.5 & 0.0 & 10.0 \\
TIES & 53.3 & 35.8 & 22.9 & 75.3 & 81.5 & 86.5 & 40.3 & 31.4 & 10.5 & 25.6 & 0.0 & 15.0 \\
DARE & 90.9 & 71.6 & 45.3 & 92.2 & 95.1 & 97.5 & 59.0 & 60.8 & 34.8 & 55.7 & \textbf{13.3} & 42.5 \\
SLERP & 91.5 & 72.2 & 46.0 & 92.2 & 94.9 & 98.0 & 59.8 & 62.3 & 36.4 & \textbf{58.1} & \textbf{13.3} & \textbf{47.5} \\
MoD (Our) & \textbf{92.4} & \textbf{75.4} & \textbf{47.0} & \textbf{94.5} & \textbf{95.4} & \textbf{98.1} & \textbf{60.6} & \textbf{64.2} & \textbf{37.6} & 51.0 & \textbf{13.3} & \textbf{47.5} \\
\hline
\end{tabular}
\end{table*}

We compare the performance of our MoD method with several established model-merging techniques, including Linear \citep{NEURIPS2022_70c26937}, Task-Arithmetic \citep{ilharco2023editing}, TIES \citep{yadav2023tiesmerging}, DARE \citep{yu2024language}, and SLERP \citep{shoemake1985animating}. These methods represent widely used and advanced approaches for merging large language models, implemented using the Mergekit package \citep{goddard2024arceesmergekittoolkitmerging}\footnote{\url{https://github.com/arcee-ai/mergekit}}.

\paragraph{Experimental Evaluation} \label{Evaluations}
We conducted extensive experiments to evaluate the effectiveness of MoD by merging two variants of Large Language Models (LLMs): Qwen-2.5 Instruct and Qwen-2.5 Math Instruct, each available in 1.5B and 7B parameter versions. The general-purpose Qwen-2.5 Instruct model serves as the base model with a density of 0.9, while the mathematics-specialized Qwen-2.5 Math Instruct contributes with a density of 0.1 across all experimental configurations (Detail configuration in Appendix \ref{ap:configuration}). This combination was specifically chosen to demonstrate MoD's capability in merging models with complementary task-specific strengths.

Our evaluation results for the 1.5B parameter models, presented in Table~\ref{tab:method-comparison-1.5B}, demonstrate MoD's superior performance across all benchmarks. On fundamental mathematical tasks such as GSM8K, MoD achieves 74.5\% accuracy, surpassing the previous state-of-the-art method DARE by a substantial margin of 23 percentage points. The performance differential becomes even more pronounced on complex benchmarks like MATH, where MoD attains 55.8\% accuracy compared to Task-Arithmetic's 27.9\%. Notably, MoD exhibits robust performance on both elementary and advanced mathematical reasoning tasks, achieving 95.1\% on MAWPS and 88.0\% on ASDiv. The method's generalization capabilities are further evidenced by strong performance on specialized benchmarks, including CARP En (56.0\%) and the challenging Olympiad Bench (20.6\%). In contrast, baseline methods including Linear, Task-Arithmetic, TIES, and SLERP demonstrate significant limitations, particularly on competitive mathematics benchmarks, with several methods failing to achieve measurable performance on AIME24, and TIES showing 0\% accuracy on AMC23.

The results for the 7B parameter models, detailed in Table~\ref{tab:method-comparison-7B}, further validate MoD's effectiveness across diverse mathematical tasks. MoD establishes new state-of-the-art benchmarks on fundamental tests, achieving 92.4\% on GSM8K and 75.4\% on MATH. This superior performance extends to practical applications, with exceptional results on MAWPS (98.1\%) and ASDiv (95.4\%). The method demonstrates particular strength in specialized domains, achieving 64.2\% on GaoKao 2023 En and 60.6\% on CARP En, substantially outperforming established methods such as SLERP and DARE. MoD's capability in advanced mathematical reasoning is further demonstrated by its leading performance on Olympiad Bench (37.6\%). While maintaining competitive performance on standardized tests (AIME24: 13.3\%, AMC23: 47.5\%), MoD's consistent superiority across varied mathematical tasks underscores its robust architecture and strong generalization capabilities.

\section{Conclusions} \label{Conclusion}
In this paper, we introduced Mixture of Distributions (MoD), a novel approach for merging Large Language Models that preserves and leverages the strengths of constituent models through probabilistic distribution combination. Our method demonstrates significant advantages over existing parameter-merging techniques by maintaining critical density characteristics while enabling selective integration of model capabilities. The experimental results across diverse mathematical benchmarks validate MoD's effectiveness, achieving state-of-the-art performance on both fundamental and advanced tasks. Our findings suggest that distribution-based merging approaches offer a promising direction for developing more capable and adaptable language models, particularly in specialized domains requiring precise knowledge integration.

\section{Limitations and Future Work} \label{Limitations-and-Future-Work}
While MoD demonstrates superior performance compared to existing methods, we acknowledge some limitations in our current study. First, our experimental validation is primarily confined to the mathematical domain, which, while comprehensive, may not fully represent the method's generalizability across other specialized fields. Second, our current approach employs a simplified strategy for determining mixture weights, which may not capture optimal combinations for all scenarios.

These limitations suggest several promising directions for future research. First, extending the evaluation of MoD to diverse domains beyond mathematics would provide valuable insights into the method's robustness and general applicability. Second, developing more sophisticated approaches for determining optimal mixture weights could potentially enhance the method's performance further. Additionally, investigating the theoretical foundations of distribution-based merging approaches could lead to more principled strategies for model combination and integration. These directions would contribute to a deeper understanding of model merging techniques and their applications in developing more capable language models.

\bibliography{anthology,custom}

\clearpage 

\onecolumn 

\appendix
\input{appendix_config}

\end{document}

%% file: appendix_config.tex
\section{YAML Configuration} \label{ap:configuration}

This appendix details the configuration parameters implemented across all methodologies in this study, specifically for the 1.5B parameter model variant. These configurations are similar to those employed in the 7B parameter implementation.

\begin{lstlisting}[style=yamlstyle, caption={MoD Method (Our)}]
base_model: Qwen/Qwen2.5-1.5B-Instruct

experts:
  - source_model: Qwen/Qwen2.5-1.5B-Instruct
  - source_model: Qwen/Qwen2.5-Math-1.5B-Instruct

model_kwargs:
  - device_map: cuda
  - low_cpu_mem_usage: True
  - trust_remote_code: True

weights: [0.9, 0.1]
\end{lstlisting}

\begin{lstlisting}[style=yamlstyle, caption={Linear Method}]
models:
  - model: Qwen/Qwen2.5-1.5B-Instruct
    parameters:
      weight: 0.9
  - model: Qwen/Qwen2.5-Math-1.5B-Instruct
    parameters:
      weight: 0.1
merge_method: linear
dtype: float16
\end{lstlisting}

\begin{lstlisting}[style=yamlstyle, caption={DARE Method}]
models:
  - model: Qwen/Qwen2.5-1.5B-Instruct
    # No parameters necessary for base model
  - model: Qwen/Qwen2.5-Math-1.5B-Instruct
    parameters:
      density: 0.9
      weight: 0.1
merge_method: dare_ties
base_model: Qwen/Qwen2.5-1.5B-Instruct
parameters:
  int8_mask: true
dtype: bfloat16
\end{lstlisting}

\begin{lstlisting}[style=yamlstyle, caption={TIES Method}]
models:
  - model: Qwen/Qwen2.5-1.5B-Instruct
  - model: Qwen/Qwen2.5-Math-1.5B-Instruct
    parameters:
      density: 0.9
      weight: 0.1
merge_method: ties
base_model: Qwen/Qwen2.5-1.5B-Instruct
parameters:
  normalize: true
dtype: bfloat16
\end{lstlisting}

\newpage
\begin{lstlisting}[style=yamlstyle, caption={Task Arithmetic Method}]
models:
  - model: Qwen/Qwen2.5-1.5B-Instruct
    parameters:
      weight: 0.9

  - model: Qwen/Qwen2.5-Math-1.5B-Instruct
    parameters:
      weight: 0.1

base_model: Qwen/Qwen2.5-1.5B-Instruct
merge_method: task_arithmetic
parameters:
  normalize: true
  int8_mask: true

dtype: bfloat16\end{lstlisting}

\begin{lstlisting}[style=yamlstyle, caption={SLERP Method}]
slices:
  - sources:
      - model: Qwen/Qwen2.5-1.5B-Instruct
        layer_range: [0, 28]
      - model: Qwen/Qwen2.5-Math-1.5B-Instruct
        layer_range: [0, 28]
merge_method: slerp
base_model: Qwen/Qwen2.5-1.5B-Instruct
parameters:
  t:
    - filter: self_attn
      value: 0.1
    - filter: mlp
      value: 0.1
    - value: 0.1
dtype: bfloat16\end{lstlisting}